\documentclass[runningheads]{llncs}
\usepackage[T1]{fontenc}
\usepackage{graphicx}
\usepackage[colorlinks]{hyperref}

\usepackage{cite}
\usepackage{bbding}  
\usepackage{pifont}  
\begin{document}
\title{SemiHMER:  Semi-supervised Handwritten Mathematical Expression Recognition using pseudo-labels}
%
%
\author{Kehua Chen\inst{1}\orcidID{0000-0003-2333-7450} \and
Haoyang Shen\inst{1}\orcidID{1111-2222-3333-4444}}
\authorrunning{K. Chen et al.}

\institute{ 
\email{\{chenkehua880, shenhaoyang6\}@gmail.com}}
%
\titlerunning{Semi-supervised HMER using pseudo-labels}

\maketitle              
\begin{abstract}
In this paper, we study semi-supervised Handwritten Mathematical Expression Recognition (HMER) via exploring both labeled data and extra unlabeled data. We propose a novel consistency regularization framework, termed SemiHMER, which introduces dual-branch semi-supervised learning. Specifically, we enforce consistency between the two networks for the same input image. The pseudo-label, generated by one perturbed recognition network, is utilized to supervise the other network using the standard cross-entropy loss. The SemiHMER consistency encourages high similarity between the predictions of the two perturbed networks for the same input image and expands the training data by leveraging unlabeled data with pseudo-labels. We further introduce a weak-to-strong strategy by applying different levels of augmentation to each branch, effectively expanding the training data and enhancing the quality of network training. Additionally, we propose a novel module, the Global Dynamic Counting Module (GDCM), to enhance the performance of the HMER decoder by alleviating recognition inaccuracies in long-distance formula recognition and reducing the occurrence of repeated characters. The experimental results demonstrate that our work achieves significant performance improvements, with an average accuracy increase of 5.47\% on CROHME14, 4.87\% on CROHME16, and 5.25\% on CROHME19, compared to our baselines.

\keywords{Handwritten Mathematical Expression Recognition \and Semi-supervised Learning \and Global Dynamic Counting Module.}
\end{abstract}
\section{Introduction} \label{sec1}
Handwritten mathematical expression recognition (HMER) is a crucial task in Optical Character Recognition (OCR) with applications in digital libraries, office automation, automatic grading. Despite significant progress, HMER remains challenging due to the complex structure of formulas, personalized handwriting styles, and the substantial cost and time involved in manual labeling. Semi-supervised learning, which leverages both labeled and unlabeled data, offers a promising solution to these challenges.

In semi-supervised learning, consistency regularization and self-training are two widely used techniques\cite{pelaez2023survey}. Consistency regularization-based methods promote the network to generate consistent predictions for the same input with different augmentations by calculating the differences between outputs as a loss function. This technique effectively utilizes unlabeled data through three main approaches: input perturbation (e.g., applying transformations like rotation or scaling to the input image), feature perturbation (e.g., introducing noise or variations in the feature space), and network perturbation (e.g., using different network architectures or parameter initializations to generate diverse predictions). 

Self-training has also been employed in semi-supervised learning research. It starts by training an initial model on labeled images to obtain pseudo-labels for unlabeled images, incorporating them to expand the training data, and then retraining the model to get boosted. One approach is called cross pseudo-supervision, which feeds both labeled and unlabeled images into two networks that share the same structure but have different parameter initializations, adding constraints to ensure similar outputs from both networks for the same sample. Here, we propose a more novel and simpler consistency regularization method where we apply pseudo-supervision on unlabeled images, which serves as an additional signal to supervise another branch network. 

Motivated by FixMatch's approach\cite{sohn2020fixmatch} of imposing consistency constraints between predictions generated from weak and strong augmentations, we develop a weak-to-strong cross-head collaborative training framework for HMER tasks, achieving higher accuracy than conventional supervised training structures that rely solely on labeled data. In our framework, weak augmentation refers to applying minimal even no image transformation, while strong augmentation involves applying diverse and extensive transformations (e.g., distortion, stretch, perspective). Our weak-to-strong augmentation strategy applies weak augmentation to one branch and strong augmentation to the other, intentionally creating a significant divergence in their predictions. By enforcing consistency regularization between these two branches, we ensure that the model learns robust representations from the same input image under different augmentation conditions. This approach effectively expands the training data, leading to enhanced training quality of the recognition network. 

Meanwhile, we propose a GDCM (Global Dynamic Counting Module) to alleviate the misrecognition issues in long-distance formulas and the misrecognition of repeated identical characters. While updating the hidden state and context information at each step, GDCM dynamically updates the obtained global counting vector based on the prediction from the previous time step. The updated counting vector is then used to compute the output probability at the current time step. 
The main contributions of this paper can be summarized as follows.
\begin{enumerate}
    \item To the best of our knowledge, we are the first to apply semi-supervised pseudo-supervision learning to the HMER task, which can further improve generalization ability and learn more compact feature representations from two different perspectives, using pseudo-labels.
    \item We introduce an innovative weak-to-strong augmentation strategy designed to enhance the performance of pseudo-supervision learning. By progressively applying stronger image augmentation techniques, our approach not only improves model robustness but also maximizes the utilization of training data, leading to significant advancements in recognition accuracy. 
    \item We propose a novel module GDCM to enhance the performance of the HMER decoder, which alleviates recognition inaccuracy issues in long-distance formula recognition and repeated character occurrences.
\end{enumerate}
\section{Related work} \label{sec2}
\subsection{HMER} \label{sec2.1}
Handwritten mathematical expression recognition (HMER) in the field of optical character recognition (OCR) faces significant challenges due to the diverse writing styles and the nested, hierarchical structures of handwritten expressions. Traditional methods\cite{chan1998elastic, hu2011hmm, keshari2007hybrid, kosmala1999line, vuong2010towards, winkler1996hmm} typically involve two steps: first identifying individual characters and then applying grammatical rules for correction. However, Limited feature-learning capabilities and complex grammatical rules make these traditional methods inadequate for practical applications. With the widespread application and impressive performance of machine learning and deep learning networks in academia, encoder-decoder architectures have increasingly been applied to handwritten formula recognition. This end-to-end recognition framework, resembling a global recognition approach, has significantly improved accuracy in handwritten expression recognition and effectively handles the strong contextual dependencies within formulas. 

Current deep learning-based formula recognition methods can be categorized into two types based on the decoding strategy: one method decodes the formula as a LaTeX expression string, while the other employs tree decoding based on the positional relationships between parent and child nodes. Both approaches have surpassed traditional formula recognition methods in terms of accuracy.

\begin{itemize}
    \item [•] 
    \textbf{Sequence-based Methods.}
    The sequence-based decoding approach for handwritten formula recognition was first introduced by Deng et al.\cite{awal2014global}, achieving the initial transformation from image to LaTeX expression via an encoder-decoder architecture. Building on this, Zhang et al.\cite{zhang2017watch} proposed the WAP model which utilizes a fully convolutional encoder to extract image features and introduces a coverage attention mechanism to address attention inaccuracies in long-distance decoding, establishing WAP as the foundational model for most sequence-based recognition methods. Zhang et al.\cite{zhang2018multi} further improved upon this with DenseWAP, which employs a DenseNet structure to enhance the encoder and proposes a multi-scale attention decoding model to address challenges in recognizing characters of varying sizes within formulas. In recent years, numerous advancements\cite{zhang2017gru, zhang2018track, li2020improving, wu2019image, wu2020handwritten, guo2022primitive, lin2022cclsl, truong2020improvement, li2022counting, bian2022handwritten, liu2023semantic} have been built on this model. Wu et al.\cite{wu2019image, wu2020handwritten} introduced adversarial learning with handwritten and printed text, while Guo et al.\cite{guo2022primitive} and Ling et al.\cite{lin2022cclsl} incorporated contrastive learning, aiding the model in learning style-invariant semantic features and reducing the influence of handwriting styles on formula recognition. Truong et al.\cite{truong2020improvement} and Li et al.\cite{li2022counting} incorporated weak supervision modules into the encoder-decoder framework to provide additional feature information, enhancing model performance. Bian et al.\cite{bian2022handwritten} introduced mutual learning to handwritten formula recognition, leveraging complementary information to alleviate challenges in long-distance decoding. SAM\cite{liu2023semantic} builds a semantic graph based on statistical co-occurrence probabilities, which explicitly demonstrates the dependencies between different symbols; it also proposes a semantic-aware module that takes visual and classification features as inputs and maps them into the semantic space. 
\end{itemize}
\begin{itemize}
    \item [•] 
    \textbf{Tree-based Methods.}
    Tree-based methods convert data expressions into tree structures, which allows for hierarchical relationships to be modeled and enables tree decoders to learn features grounded in grammatical rules. Zhang et al.\cite{zhang2020tree} proposed the first image-to-token tree decoding model TDv1. Building on this, Wu et al.\cite{wu2022tdv2} introduced the TDv2 model, which removes prioritization among branches of tree nodes to enhance the model’s generalization capability and employs a diversified tree structure labeling system to greatly simplify the decoding process. Yuan et al.\cite{yuan2022syntax} presented the first tree decoding model that integrates syntactic information into the encoder-decoder architecture, introducing decomposition rules based on the syntactic characteristics of formulas. This approach segments formulas into different components to construct the formula tree, effectively mitigating ambiguities associated with tree structures. TAMER\cite{zhu2024tamer} integrates the strengths of both sequence and tree decoding models, enhancing the model's comprehension and generalization of complex mathematical expression structures through the simultaneous optimization of sequence prediction and tree structure prediction objectives. PosFormer\cite{guan2025posformer} jointly optimizes expression recognition and position recognition tasks to explicitly enable position-aware symbol feature learning for representation.
\end{itemize}

\subsection{Semi-supervised learning} \label{sec2.2}
SSL(Semi-supervised learning) is an effective approach to improve model recognition accuracy in scenarios where there is a small amount of labeled data and a larger amount of unlabeled data. These algorithms can be broadly divided into the following categories:

\begin{itemize}
    \item [•] 
    \textbf{Consistency training.}
    Based on the assumption that if a reasonable perturbation is applied to an unlabeled dataset, the model's prediction should not change significantly. Therefore, the model can be trained to produce consistent predictions on a given unlabeled sample and its perturbed version.
\end{itemize}

\begin{itemize}
    \item [•]
    \textbf{Pseudo-label Method.}
    Pseudo-labeling or self-training is a typical technique for leveraging unlabeled data. It alternates between pseudo-label prediction and feature learning, encouraging the model to make confident predictions for unlabeled data. 
\end{itemize}

Semi-supervised learning has been extensively applied in various areas such as image classification, object detection, and image segmentation.

\textbf{Semi-supervised classification}. 
UDA\cite{xie2020unsupervised} employs a back-translation data augmentation method and then utilizes semi-supervised data augmentation techniques in classification tasks, thereby improving classification accuracy. Yalniz et al.\cite{yalniz2019billion} adopts a teacher/student learning mechanism and leveraging a billion-level amount of unlabeled data along with a relatively small quantity of labeled data, the performance of existing models in image classification tasks has been improved. GLA-DA\cite{tu2024gla} proposed a domain adaptation method for multivariate time series data that leverages global and local feature alignment to improve classification performance in both unsupervised and semi-supervised settings. 

\textbf{Semi-supervised segmentation}.
ST++\cite{yang2022st++} method introduces a reliable pseudo-label selection mechanism along with a weak-to-strong consistency strategy. This allows the model to more accurately identify and leverage high-quality pseudo-labels, reducing the negative impact of noisy data on the segmentation model. MixMatch\cite{berthelot2019mixmatch}, combines data augmentation, consistency regularization, and entropy minimization. It mixes labeled and unlabeled data while applying data augmentation, which helps the model learn more effectively from limited labeled data.  CPS\cite{chen2021semi} uses two networks with identical structures but different initialization. It adds a constraint such that the one-hot pseudo-label generated by one network for a given sample is used as the target for the other network's prediction, with the process supervised by cross-entropy loss. 

\textbf{Semi-supervised object detection}.
STAC\cite{sohn2020simple} proposed a semi-supervised object detection algorithm based on hard pseudo-labels. Ambiguity-Resistant Semi-supervised Learning (ARSL)\cite{liu2023ambiguity}, including Joint-Confidence Estimation (JCE) and Task-Separation Assignment (TSA), is generally applicable to single-stage semi-supervised object detection tasks. ISTM\cite{yang2021interactive} proposed an interactive self-training model. It integrates object detection results from different iterations using Non-Maximum Suppression (NMS) and uses two ROI heads with different structures to estimate each other's pseudo-labels.  TLDR\cite{wang2022consistent} proposed a new semi-supervised object detector called Consistent-Teacher by analyzing the existing issues of pseudo-label misalignment and inconsistency in semi-supervised object detection. 

Despite these advancements, there is almost no research on the application of semi-supervised learning in the field of handwritten mathematical expression recognition.  

\section{The Approach} \label{sec3}
From the perspective of architecture,  we propose typical double-branch mutual learning methods and from strong to weak data augmentation, conservative-progressive collaborative learning to obtain additional training signals. The two branches are trained simultaneously with heterogeneous knowledge to reduce misleading and model coupling problems. 

\subsection{Overview} \label{sec3.1}
\begin{figure}[htb]
    \centering
    \includegraphics[width=1\linewidth]{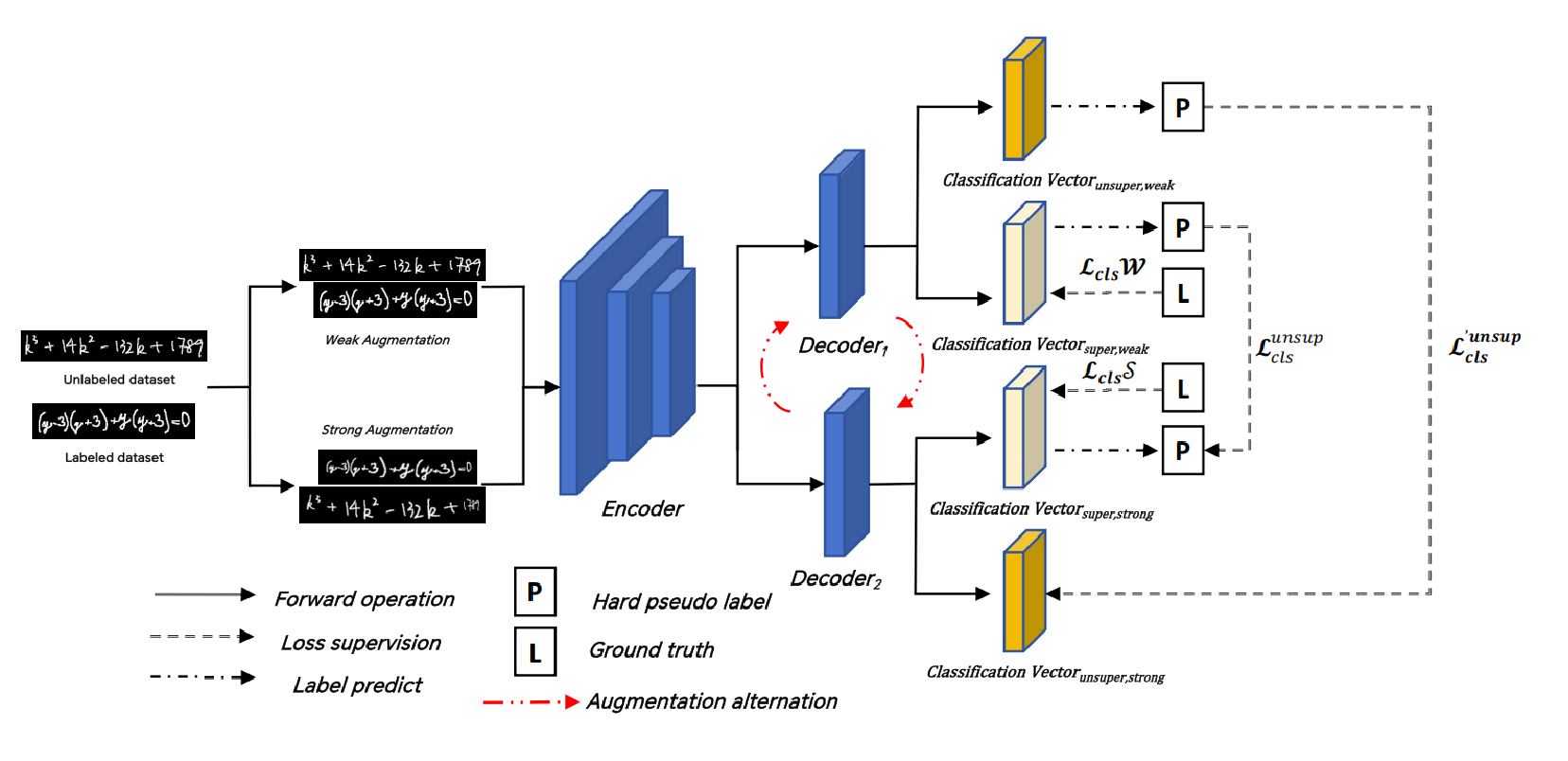}
    \caption{Overview of our Semi-supervised Handwritten Mathematical Expression Recognition using pseudo-labels. In this scheme, we use ${Decoder_1}$ as an example. Weakly augmented (non-augmented) labeled data flows through the encoder module and the corresponding prediction ${Decoder_1}$ to produce the prediction ${Classification\  Vector_{super,weak}}$, which is then supervised by the ground truth. At the same time, the prediction ${Classification\  Vector_{super,weak}}$ will be transformed into hard pseudo labels, which also utilized as a supervision signal for the strongly augmented image prediction ${Classification\  Vector_{super,strong}}$ from the other head. The ${Classification\  Vector_{super,strong}}$ is also supervised by the ground truth labels. Besides that, We use predictions ${Classification\  Vector_{unsuper,weak}}$ from weakly augmented unlabeled data as ground truth to supervise the encoder and predictions ${Classification\  Vector_{unsuper,strong}}$ on strongly augmented data. Augmentation alternation refers to alternately switching the intensity of augmentation applied to different decoders.}
    \label{fig1}
\end{figure}
Figure \ref{fig1} shows our proposed semi-supervised framework for handwritten mathematical expression recognition. The framework employs a cross-decoder architecture that processes both labeled and unlabeled data through a shared network. Following CAN\cite{li2022counting}, we adopt DenseNet\cite{huang2017densely} as the backbone for feature extraction. Instead of using separate individual models, we innovatively design two identical decoder heads working in parallel, which we term the cross-head structure. This design enables more efficient feature learning and enhances the model's generalization capability through mutual supervision.

Our semi-supervised learning strategy operates in two parallel streams. For labeled data processing, we first feed weakly augmented images through the encoder and one of the decoders to generate initial predictions. These predictions serve dual purposes: they are supervised by ground truth labels for word prediction and transformed into hard pseudo-labels to guide the other decoder's predictions on strongly augmented versions of the same images. This cross-supervision mechanism ensures consistency between different augmentation views while maintaining alignment with ground truth. For unlabeled data, we employ a similar but simplified approach: the predictions from weakly augmented images serve as pseudo-labels to supervise the predictions of their strongly augmented counterparts, effectively leveraging unlabeled data through consistency training. 

The proposed approach is composed of a shared backbone $F$ and two parallel decoders $G_m(m \in \{1,2\})$, with both decoders having identical structures. The pseudo label is generated by $P_m(y|x)=G_m(F(x))$, where $F$ is the backbone network and $Gm$ is the respective decoder. This pseudo-label is subsequently used as the supervision signal for the other decoder, creating a feedback loop that enhances learning.

To enhance the robustness of both decoders, we employed a rotational alternating training strategy. During the ${i}-th$ epoch, we apply weak augmentation to ${Decoder_1}$ while using strong augmentation for ${Decoder_2}$. Conversely, in the ${(i+1)}-th$ epoch, we switch the augmentation strategies: applying weak augmentation to ${Decoder_2}$ and strong augmentation to ${Decoder_1}$. This alternating pattern is repeated iteratively throughout the training process.  

\subsection{Weak-to-strong augmentation} \label{sec3.2}
To fully leverage the advantages of consistency training, We utilize weak and strong augmentations to introduce additional information into our framework. In our experiments, weak augmentation means no augmentation, whereas strong augmentation combines distortions, stretching, and perspective changes. Specifically, we randomly distort images every time, stretch images with a probability of 50\%, and perspective images with a probability of 30\%.

Our approach creates pseudo-labels through consistency regularization and self-training. These pseudo-labels are intermediate outputs generated from images with both weak and strong augmentations. They serve as a supervisory signal when the model processes two images simultaneously. We've devised a pool of operations that includes three kinds of image transformations. Rather than applying a uniform intensity level of transformation across all training iterations, we choose transformations for each sample in a mini-batch randomly from a set range at every training epoch. 

By alternating the application of augmentations and dynamically training the decoders, our framework not only maximizes the effective use of both labeled and unlabeled data but also fosters a robust learning environment that is capable of accurately recognizing handwritten mathematical expressions. This systematic approach ultimately leads to improved performance in real-world applications.

\subsection{Global Dynamic Counting Module} \label{sec3.4}
\begin{figure}[htb]
    \centering
    \includegraphics[width=0.85\linewidth]{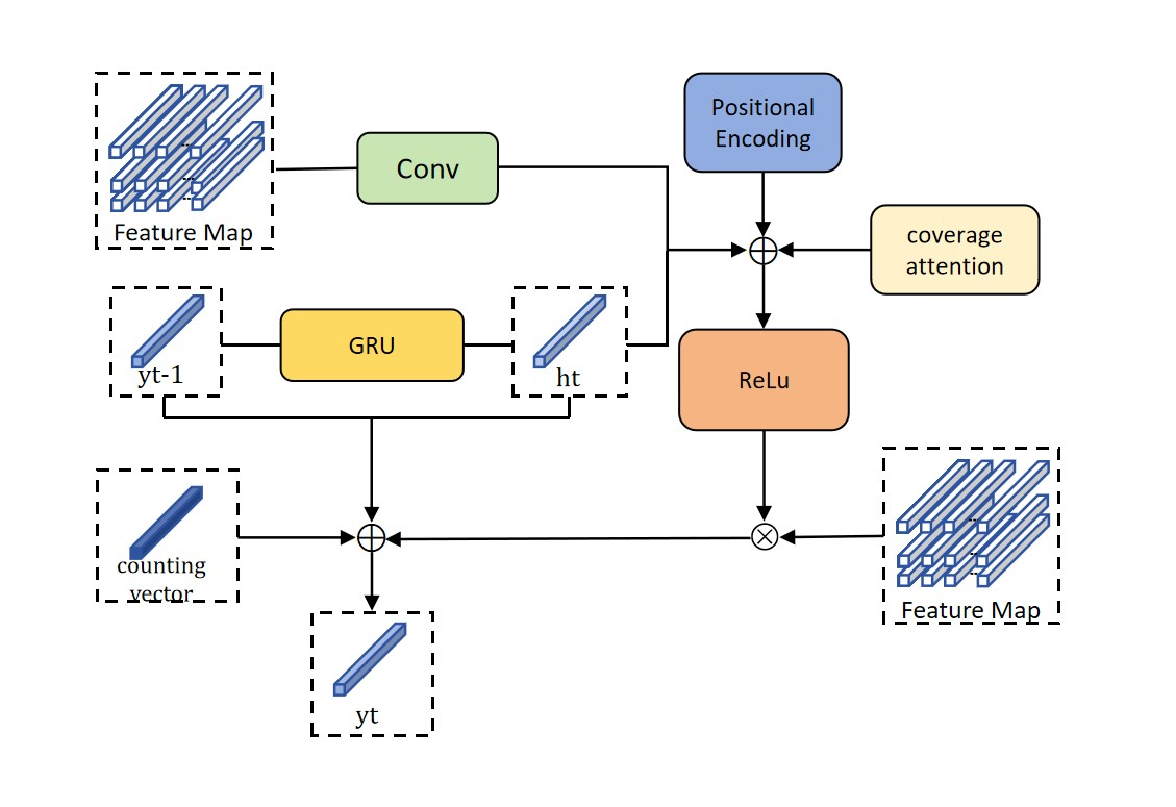}
    \caption{Structure of the proposed decoder based on the Global Dynamic Counting Module (GDCM)}
    \label{fig2}
\end{figure}

CAN\cite{li2022counting} calculates the global counting vector from the global image features by the encoder using an additional multi-scale counting module. This global counting vector is then fed into the decoder to inform the decoding process. By providing the decoder with this additional weak supervision information, the model's performance is effectively improved. However, in this approach, the global count vector is only utilized as auxiliary information to calculate the current output, without being more deeply integrated with the formula prediction task. As a result, it cannot provide the model with more useful guidance during decoding, and the issues of long-distance formula recognition problems and errors when the same character appears repeatedly remain challenging to address.

To overcome these limitations and better leverage the relationship between formulas and count vectors to guide the model, we introduce the concept of \textbf{GDCM}(Global Dynamic Counting Module) into the Handwritten Mathematical Expression Recognition task. Specifically, based on the encoder-decoder recognition logic, the decoder relies on the previous state to predict the current output at each time step. Therefore, the feature information input to the decoder should be dynamically updated at each step. Our model achieves this by dynamically updating the global count vector based on the predicted value from the previous time step, while also updating the hidden state and context information. The updated count vector is then used to calculate the output probability for the current time step.

In their work, the decoder uses the context vector C, the hidden state ht, and the embedding $E(y_{t-1})$ to predict $y_t$. To address the lack of global information, they introduce the counting vector V, which is calculated from a global counting perspective, as additional global information to make the prediction more accurate. The prediction of $y_t$ is combined as follows:

\begin{equation}
    p(y_t) = softmax(w_oT (W_c C + W_v V + W_t h_t + W_e E) + b_o) 
    \label{Eq1}
\end{equation}
\begin{equation}
    y_t \sim p(y_t)
    \label{Eq2}
\end{equation}

Then, we use dynamic count vectors to update the predictions again, as shown below:

\begin{equation}
    p(\widehat{y}_t) = W_k \cdot (\widehat{h}_t - \sum_{i=1}^{C} softmax(p(y_t) )) + b_k
    \label{Eq3}
\end{equation}

where $w_o $,$b_o$,$W_c$,$W_v$,$W_t$,$W_e$,$W_k$,$b_k$ are all trainable weights.

\subsection{Loss function} \label{sec3.5}
In the setting of semi-supervised learning, we provide a batch of labeled samples $\mathcal{D}_l$ at each iteration. We preliminarily $\mathcal{L}_{cls}$ as standard cross entropy classification loss of the word predicted probability $\mathcal{P}(y_t)$, with $\mathcal{W}$ and $\mathcal{S}$ representing the corresponding weak and strong augmentation functions applied to the image respectively. The $\mathcal{L}_{cls}^{unsup}$ defined as:

\begin{equation}
    \mathcal{L}_{cls}^{unsup} = \mathcal{L}_{cls}(\mathcal{P}^\mathcal{S}(y_{t}) ,\mathcal{P}^\mathcal{W}(y_{t}))
    \label{Eq4}
\end{equation}                              
where $\mathcal{L}_{cls}^{sup}$ was defined as :

\begin{equation}
    \mathcal{L}_{cls}^{sup} = \mathcal{L}_{cls} \mathcal{W}(x_{t}) + \mathcal{L}_{cls} \mathcal{S}(x_{t})
    \label{Eq5}
\end{equation}      
For the unlabeled samples $\mathcal{D}_{un}$ provided at each iteration, we also use the standard cross-entropy classification loss for supervision. Specifically, we denoted the unlabeled samples's predictions as $y_{t}^{'}$. The predictions from weak supervision are used to supervise the predictions from strong supervision, defined as $\mathcal{L}_{cls}^{unsup}$ as follows: 
\begin{equation}
    \mathcal{L^{'}}_{cls}^{unsup} = \mathcal{L}_{cls}^{unsup}(\mathcal{P}^\mathcal{S}(y_{t}^{'}) ,\mathcal{P}^\mathcal{W}(y_{t}^{'}))
    \label{Eq6}
\end{equation}    
Furthermore, for each decoder, there is a counting ground truth loss for each symbol class of decoder's prediction result, represented by $\widehat{\mathcal{V}}$. It is a smooth L1 loss\cite{ren2015faster} function used to calculate the loss between the predicted count and the ground truth for each formula (not shown in \ref{fig1}), defined as follows:

\begin{equation}
    \mathcal{L}_{counting} = Smooth\mathcal{L}_{1} \cdot \mathcal{W}(\widehat{\mathcal{V}}_{1}) + Smooth\mathcal{L}_{1} \cdot \mathcal{S}(\widehat{\mathcal{V}}_{2})
    \label{Eq7}
\end{equation}
                                    
So the total loss function is composed of four components and is defined in the following manner:
\begin{equation}
    \mathcal{L} =  \mathcal{L}_{cls}^{unsup} +  
    \mathcal{L}_{cls}^{sup} +
    \mathcal{L^{'}}_{cls}^{unsup} +
    \mathcal{L}_{counting} 
    \label{Eq8}
\end{equation}

\section{Experiments} \label{sec4}
\subsection{Datasets} \label{sec4.1}

\begin{figure}[htb]
    \includegraphics[width=\textwidth]{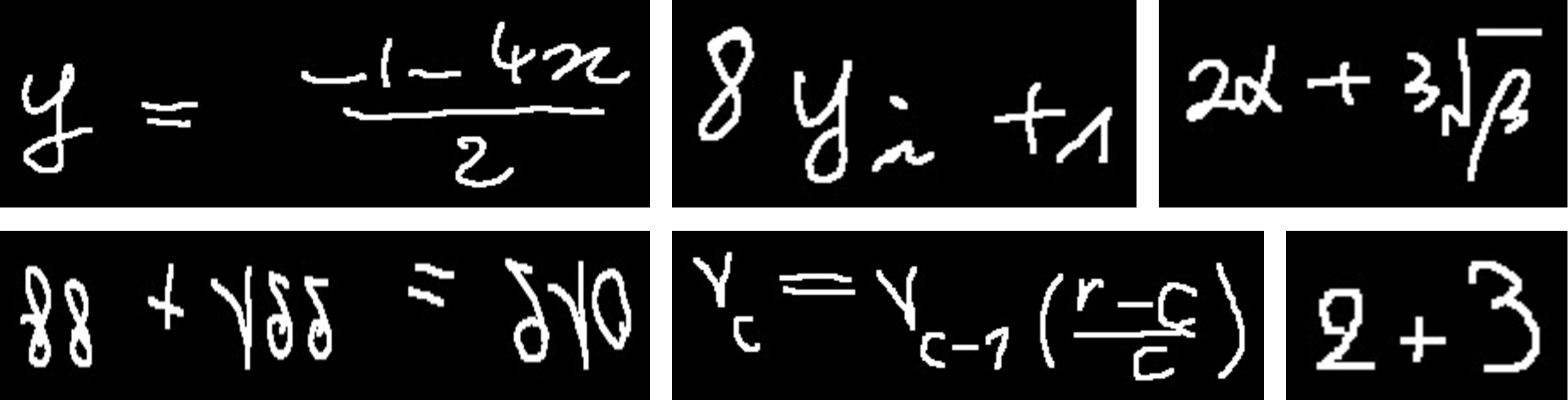}
    \caption{Some example images from the CROHME dataset.}
    \label{fig3}
\end{figure}

We conduct experiments on CROHME benchmark datasets and compare the performance with the state-of-the-art methods. 

\textbf{CROHME Dataset}\cite{mouchere2014icfhr} is from the competition on recognition of online handwritten mathematical expression, which is the most widely used public dataset. It is derived from the competition focused on recognizing online handwritten mathematical expressions. The CROHME dataset creates images from handwritten stroke trajectory data found in InkML files, leading to images with clean backgrounds. The CROHME training set number is 8,836, while the test set contains 986, 1147 and 1199 images, including 111 symbol classes (C), including "sos" and "eos". Some example images are shown in Fig. \ref{fig2}. We randomly selected 1,000 images from the CROHME 2023 dataset\cite{xie2023icdar} as unlabeled data, with all characters included in the CROHME Dataset.

\subsection{Implementation details} \label{sec4.2}
The proposed SemiHMER is implemented in PyTorch. To accelerate the iteration efficiency, we used four Nvidia Tesla V100 GPUs, each equipped with 32GB RAM, to train the model. It is worth noting that training SemiHMER on a single GPU would take at least one week, which is highly inefficient. The batch size is set to 12, and the Adadelta\cite{zeiler2012adadelta} optimizer is used for training. The learning rate starts at 0 and monotonically increases to 2 by the end of the first epoch. After that, the learning rate decays to 0 according to a cosine annealing schedule. To enhance the effectiveness of semi-supervised learning, we did not use cross-training in the first 240 epochs, allowing each branch to train independently. After 240 epochs, we conducted semi-supervised cross-training with a learning rate of 0.2 and further fine-tuned for another 240 epochs. Our method relies on data augmentation techniques, and we employed various augmentation methods (distortion, affine transformation, and stretching) to explore the capabilities of our approach.

\subsection{Evaluation Metrics} \label{sec4.3}
We use the expression recognition rate (ExpRate) to evaluate the performance of different approaches. ExpRate is defined as the percentage of predicted mathematical expressions that exactly match the ground truth.

\begin{table}[htb]
\centering
\caption{Expression Recognition Rate (ExpRate) performance of SemiHMER and other state-of-the-art methods on CROHME 2014, CROHME 2016, and CROHME 2019 test set.}
    \begin{tabular}{cccccccccc}
    \hline 
    Method & \multicolumn{3}{c}{CROHME 2014} & \multicolumn{3}{c}{CROHME 2016} & \multicolumn{3}{c}{CROHME 2019} \\
           & ExpRate & $\leq1$ & $\leq2$ & ExpRate & $\leq1$ & $\leq2$ & ExpRate & $\leq1$ & $\leq2$ \\
    \hline 
         \multicolumn{10}{c}{Without data augmentation} \\
    \hline    
        UPV & 37.22& 44.22 & 47.26 & - & - & - & - & - & - \\
        TOKYO & - & - & - & 43.94 & 50.91 & 53.70 & - & - & - \\
        DWAP & 50.10 & - & - & 47.50 & - & - & - & - & - \\
        DWAP-TD & 49.10 & 64.20 & 67.80 & 48.50 & 62.30 & 65.30 & 51.40 & 66.10 & 69.10 \\
        DWAP-MSA & 52.80 & 68.10 & 72.00 & 50.10 & 63.80 & 67.40 & 47.70 & 59.50 & 63.30 \\
        ABM & 56.85 & 73.73 & 81.24 & 52.92 & 69.66 & 78.73 & 53.96 & 71.06 & 78.65 \\ 
        CAN & 57.00 & 74.21 & 80.61 & 56.06 & 71.49 & 79.51 & 54.88 & 71.98 & 79.40 \\
        PosFormer & 60.45 & 77.28 & 83.68 & 60.94 & 76.72 & 83.87 & 62.22 & 79.40 & 86.57 \\
        TAMER & 61.23 & 76.77 & 83.25 & 60.26 & 76.91 & 84.08 & 61.97 & 78.97 & 85.80 \\
 \textbf{CAN(baseline)}& \textbf{55.48}& \textbf{73.73}& \textbf{80.22}& \textbf{53.36}& \textbf{70.97}& \textbf{79.51}& \textbf{53.38}& \textbf{72.31}&\textbf{79.40}\\
    \hline
         \multicolumn{10}{c}{With data augmentation}\\
    \hline
        CAN-DWAP &   65.58&77.36&83.35&  62.51& 74.63&82.48&   63.22&78.07&82.49\\
        CAN-ABM &   65.89&77.97&84.16&  63.12& 75.94&82.74&   64.47&78.73&82.99\\
        \textbf{baseline-aug}&   \textbf{58.32}&\textbf{75.05}&\textbf{81.54}&  \textbf{57.28}& \textbf{74.11}&\textbf{81.34}&   \textbf{57.63}&\textbf{75.15}&\textbf{82.24}\\
        \textbf{SemiHMER}&   \textbf{60.95(+5.47)}&\textbf{76.27}&\textbf{82.25}&  \textbf{58.23(+4.87)}& \textbf{73.93}&\textbf{81.51}&   \textbf{58.63(+5.25)}&\textbf{76.98}&\textbf{83.15}\\
    \hline
    \end{tabular}
\label{tab1}
\end{table}

\subsection{Comparison with State-of-the-Art} \label{sec4.4}
To demonstrate the superiority of SemiHMER, we compare it with previous state-of-the-art methods. Since SemiHMER heavily relies on the effectiveness of data augmentation, we primarily focus on the results obtained when data augmentation is included during the model training process. 
As shown in Table. \ref{tab1}, using CAN as the baseline, our multi-GPU training with four V100 GPUs and a learning rate of 2 did not achieve the original performance of CAN, yielding only 55.48\% on CROHME 2014, 53.36\% on CROHME 2016 and 53.38\% on CROHME 2019 which are lower than the results proposed by CAN. However, despite extensive tuning of hyperparameters and configurations, we were unable to fully close this gap, highlighting a limitation of our current implementation. After applying data augmentation, the performance improved by 2.84\%(58.32\%) on CROHME 2014, 3.84\%(57.2\%) on CROHME 2016 and 4.25\%(57.63\%) on CROHME 2019. With the aid of optimal data augmentation, dual-branch semi-supervised learning further enhanced the model by 2.63\%(60.95\%) on CROHME 2014, 1.03\%(58.23\%) on CROHME 2016 and 1.0\%(58.63\%) on CROHME 2019 indicates that our method can effectively leverage data augmentation significantly boost the model's performance.

\subsection{Ablation Study} \label{sec4.5}

In this subsection, we discuss the contribution of each component within our framework. 
\subsection{Effectiveness of Global Dynamic Counting Module} \label{sec4.6}

\begin{table}[htb]
\centering
\caption{Alabtion study of Global Dynamic Counting Module.}
    \begin{tabular}{cccc}
    \hline 
      Method&\multicolumn{3}{c}{Result} \\
             &CROHME 2014 & CROHME 2016 & CROHME 2019\\
    \hline
              no GDCM&55.48& 53.36& 53.38\\
             with GDCM&\textbf{56.30}& \textbf{56.14}&\textbf{53.54}\\
    \hline
    \end{tabular}
\label{tab2}
\end{table}
GDCM boosts performance to 56.30,56.14,53.54 by updating the hidden state and context information, which shows the effectiveness of our Global Dynamic Counting method.

\subsection{Effectiveness of the weak-to-strong augmentation strategy} \label{sec4.7}
By utilizing the weak-strong strategy, we can introduce more information to improve consistency. The strong transformation function generates augmented images with different intensities each time, which leads to improved model performance. However, excessive augmentation can cause a performance decline, possibly due to a significant increase in erroneous predictions from the other branch, misleading the network's optimization direction. To allow the model to benefit from both data augmentation and cross-pseudo supervision strategies, we conducted experiments with different augmentation intensities. A natural idea is to gradually increase the intensity of augmentation. As shown in Table \ref{tab3}, by progressively introducing warping, stretching, and perspective transformations into the model, performance on the dataset CROHME2014 improved by 0.09\%, 0.58\%, and 2.02\% respectively. Finally, when applying both weak-strong augmentations and cross-pseudo supervision strategies to both branches, the results further improved by 1.92\%, 2.13\%, and 2.63\%, demonstrating the effectiveness of our strategy: as the augmentation becomes more effective, the impact of CPS becomes more pronounced.  

\begin{table}[htb]
\centering
\caption{Ablation study of different augmentations strategy for pseudo supervision learning strategy.}
    \begin{tabular}{cccc|ccc}
    \hline 
    \multicolumn{4}{c|}{Strategy} & \multicolumn{3}{c}{Result} \\
        distort & stretch & perspective & cps & CROHME 2014 & CROHME 2016 & CROHME 2019\\
    \hline
        \ding{55}& \ding{55} & \ding{55} & \ding{55} &  56.30& 56.14& 53.54\\
        \Checkmark& \ding{55} & \ding{55} & \ding{55} & 56.39(+0.09)& 56.61(+0.47)&53.71(+0.17)\\
        \Checkmark & \ding{55} & \ding{55} & \Checkmark & 58.01(+1.92)& 56.93(+0.30)& 57.13(+3.59)\\
        \Checkmark & \Checkmark & \ding{55} & \ding{55} & 56.88(+0.58)& 56.23(+0.09)& 57.29(+3.75)\\
        \Checkmark & \Checkmark & \ding{55} & \Checkmark &  59.01(+2.13)& 56.66(+0.52)& 58.13(+4.59)\\
        \Checkmark & \Checkmark & \Checkmark & \ding{55} & 58.32(+2.02)& 57.27(+1.13)& 57.63(+4.09)\\
        \Checkmark & \Checkmark & \Checkmark & \Checkmark & \textbf{60.95(+2.63)}&  \textbf{58.15(+2.01)}& \textbf{58.63(+5.09)}\\
    \hline
    \end{tabular}
\label{tab3}
\end{table}

We ablated each component of SemiHMER step by step. \ref{tab4}, shows that directly applying weakly augmented samples to both branches results in a 1.53\% decrease in performance, which is possible because, without augmentation, both branches only introduce unhelpful noisy errors to the model. Naturally, we considered that simultaneously applying strong augmentation to both branches, enhancing their respective capabilities, would improve the credibility of the supervision signal. However, the result was only a 1.3\% gain, which might be attributed to the fact that simultaneous strong augmentation would exponentially increase the amount of erroneous supervisory information introduced by both branches. Finally, our method, which uses a strategy of weak augmentation on one branch and strong augmentation on the other branch, improved the model's performance by 4.65\%, demonstrating that the proposed weak-to-strong augmentation strategy for cross-pseudo supervision is a more powerful tool for semi-supervised handwritten mathematical expression recognition. 

\begin{table}[htb]
\centering
\tabcolsep=4pt
\caption{Ablation study of different augmentation strategies for both branches}
    \begin{tabular}{ccccccc}
    \hline  
         Method & \multicolumn{2}{c}{CROHME 2014}& \multicolumn{2}{c}{CROHME 2016}& \multicolumn{2}{c}{CROHME 2019}\\
                &  branch1&  branch2& branch1&branch2& branch1&branch2\\
    \hline  
        all-weak-augment& 54.77 &53.96& 53.71&54.75&50.71&50.95\\  
        all-strong-augment & 57.40&58.51& 53.79&55.36& 53.54&53.79\\
        weak-to-strong augment & \textbf{60.95}&\textbf{60.34}& \textbf{58.38}&\textbf{58.24}& \textbf{58.30}&\textbf{58.30}\\
    \hline 
    \end{tabular}
\label{tab4}
\end{table}

\subsection{The trade-off weight $\lambda$} \label{sec4.9}
$\lambda$ is used to balance the trade-off between supervised loss and unsupervised loss. Table \ref{tab5} shows that $\lambda=10^{-3}$ performs best in our setting, where a smaller $\lambda=10^{-4}$ reduces much of the useful information provided by pseudo-labels. A larger $\lambda=10^{-2}$ is problematic and leads to performance degradation because the network could converge in the wrong direction.  

\begin{table}[htb]
\centering
\tabcolsep=5pt
\renewcommand\arraystretch{1.2}
\caption{Performance analysis of different weight \textit{$\lambda$}.}
    \begin{tabular}{c|ccc}
    \hline 
        $\lambda$ & $10^{-2}$ & $10^{-3}$ & $10^{-4}$\\
    \hline 
        ExpRate & 33.6 & \textbf{60.95} & 58.22 \\
    \hline
    \end{tabular}
\label{tab5}
\end{table}

\section{Conclusion} \label{sec5}
In this paper, we propose a novel framework called SemiHMER for the field of handwritten mathematical expression recognition. Our method is the first to merge weak and strong augmentations into the cross-head co-training framework, which naturally combines the benefits of consistency and self-training. On one hand, our proposed SemiHMER boosts the diversity of consistency training samples, addressing the challenge of mutual learning between supervision signals generated by two branches. On the other hand, our method enhances the learning effect of self-training pseudo-labels through strong augmentation and weak augmentation. We demonstrate the effectiveness of our paradigm in semi-supervised handwritten mathematical expression recognition, focusing on two commonly used benchmarks CROHME. 

Although handwritten mathematical expression recognition has been well developed over the past few decades, the effectiveness of consistency regularization and self-training has not received much attention. Therefore, how to exploit the potential benefits of noisy pseudo-labels while utilizing semi-supervised learning is of great significance for future research. 

%
\bibliographystyle{splncs04}
\bibliography{citation}
\end{document}